\documentclass[alpha-refs]{ijcrms_arxiv}

\usepackage{graphicx}
\usepackage[space]{grffile}
\usepackage{latexsym}
\usepackage{textcomp}
\usepackage{longtable}
\usepackage{tabulary}
\usepackage{booktabs,array,multirow}
\usepackage{amsfonts,amsmath,amssymb}
\usepackage{natbib}
\usepackage{url}
\usepackage{hyperref}
\hypersetup{colorlinks=false,pdfborder={0 0 0}}
\usepackage{etoolbox}
\usepackage{lineno}
\usepackage[normalem]{ulem}
\usepackage{float}

\AtBeginDocument{\DeclareGraphicsExtensions{.pdf,.PDF,.eps,.EPS,.png,.PNG,.tif,.TIF,.jpg,.JPG,.jpeg,.JPEG}}

\usepackage[utf8]{inputenc}
\usepackage[english]{babel}

\usepackage[nolist,nohyperlinks]{acronym}
\usepackage{moreverb}
\usepackage{xprintlen}
\usepackage{multirow}
\usepackage{hhline}
\usepackage{booktabs}

\newcommand\BibTeX{{\rmfamily B\kern-.05em \textsc{i\kern-.025em b}\kern-.08em
T\kern-.1667em\lower.7ex\hbox{E}\kern-.125emX}}

\newcommand{\RemoveSpaces}[1]{%
  \begingroup
  \spaceskip=1sp
  \xspaceskip=1sp
  #1%
  \endgroup}
\newcommand{\refsec}[1]{Sec.~\ref{#1}}
\newcommand{\reffig}[1]{Fig.~\ref{#1}}
\newcommand{\reftab}[1]{Tab.~\ref{#1}}

\newcommand{\refeqnns}[1]{Eqs.~\ref{#1}}
\newcommand{\refapp}[1]{Appendix~\ref{#1}}
\newcommand{\Iwr}{I_{\rm wr}}
\newcommand{\IwrS}{I_{\rm wr}^{\rm SEAS5}}
\newcommand{\IwrP}{I_{\rm wr}^\prime}
\newcommand{\nreal}{50~}
\newcommand{\trainyearstart}{1940~}
\newcommand{\trainyearend}{2005~}
\newcommand{\testyearstart}{2011~}
\newcommand{\testyearend}{2024~}
\newcommand{\valstart}{January~2006~}
\newcommand{\valend}{December~2010~}
\newcommand{\nseeds}{six~}
\newcommand{\nparams}{$\approx30$k~}

\newacro{WR}{Weather regimes}
\newacro{PC}{principal component}
\newacro{EOF}{empirical orthogonal function}
\newacro{MAE}{mean absolute error}
\newacro{MSE}{mean squared error}
\newacro{ACC}{anomaly correlation coefficient}
\newacro{CE}{coefficient of efficiency}
\newacro{MARE}{mean absolute relative error}
\newacro{RMSE}{root-mean-square error}
\newacro{CRPS}{continuous ranked probability score}
\newacro{SSR}{spread--skill ratio}
\newacro{SST}{sea surface temperature}
\newacro{SLP}{sea level pressure}
\newacro{GPH}{geopotential height}
\newacro{ENSO}{El Niño Southern Oscillation}
\newacro{NAO}{North Atlantic Oscillation}
\newacro{AO}{Artic Oscillation}
\newacro{NWP}{numerical weather prediction}
\newacro{AI}{Artificial Intelligence}
\newacro{NN}{Neural Networks}
\newacro{CDS}{Copernicus Data Store}
\newacro{ECMWF}{European Center for Medium-Range Weather Forecasts}
\newacro{AT}{Atlantic Trough}
\newacro{ZO}{Zonal Regime}
\newacro{ST}{Scandinavian Trough}
\newacro{AR}{Atlantic Ridge}
\newacro{EuBL}{European Blocking}
\newacro{ScBL}{Scandinavian Blocking}
\newacro{GL}{Greenland Blocking}
\newacro{DTM}{digital terrain model}
\newacro{EQM}{empirical quantile mapping}

\definecolor{cadmiumgreen}{rgb}{0.0, 0.42, 0.24}
\definecolor{steelblue}{rgb}{0, 0.467, 0.714}
\definecolor{violet}{rgb}{0.616, 0.306, 0.867}
\definecolor{orange}{rgb}{0.918, 0.451, 0.09}

\papertype{Research Article}

\title{AI reconstruction of European weather from the Euro-Atlantic regimes}

\author[1]{Alessandro Camilletti}
\author[1]{Gabriele Franch}
\author[1]{Elena Tomasi}
\author[1]{Marco Cristoforetti}

\affil[1]{Data Science for Industry and Physics, Fondazione Bruno Kessler, via Sommarive 18, 38123, Trento (TN), Italy}

\runningauthor{A. Camilletti}
\corraddress{Alessandro Camilletti, Data Science for Industry and Physics, Fondazione Bruno Kessler, via Sommarive 18, 38123, Trento (TN), Italy}
\corremail{acamilletti@fbk.eu}
\fundinginfo{ICSC–Centro Nazionale di Ricerca in High Performance Computing, Big Data and Quantum Computing, Spoke4 - Earth and Climate,
funded by European Union – NextGenerationEU.}

\begin{document}

\maketitle 

\selectlanguage{english}
\begin{abstract}
We present a non-linear AI model designed to reconstruct monthly mean anomalies of the European temperature and precipitation based on the Euro-Atlantic \ac{WR} indices.
\ac{WR} represent recurrent, quasi-stationary, and persistent states of the atmospheric circulation that exert considerable influence over the European weather, therefore offering an opportunity for sub-seasonal to seasonal forecasting. While much research has focused on studying the correlation and impacts of the \ac{WR} on European weather, the estimation of ground-level meteorological variables, such as temperature and precipitation, from Euro-Atlantic \ac{WR} remains largely unexplored and is currently limited to linear methods.
The presented AI model can capture and introduce complex non-linearities in the relation between the \ac{WR} indices, describing the state of the Euro-Atlantic atmospheric circulation and the corresponding surface temperature and precipitation anomalies in Europe. 
We discuss the AI-model performance in reconstructing the monthly mean two-meter temperature and total precipitation anomalies in the European winter and summer, also varying the number of \ac{WR} used to describe the monthly atmospheric circulation.
We assess the impact of errors on the \ac{WR} indices in the reconstruction and show that a mean absolute relative error below 80\% yields improved seasonal reconstruction compared to the ECMWF operational seasonal forecast system, SEAS5.
As a demonstration of practical applicability, we evaluate the model using \ac{WR} indices predicted by SEAS5, finding slightly better or comparable skill relative to the SEAS5 forecast itself.
Our findings demonstrate that \ac{WR}-based anomaly reconstruction, powered by AI tools, offers a promising pathway for sub-seasonal and seasonal forecasting.

\end{abstract}

\twocolumn

\section{Introduction}
\label{sec:introduction}
Predicting atmospheric conditions for the upcoming season is crucial in various social and economic applications. Accurate seasonal forecasts can guide agricultural planning, allowing farmers to optimize crop selection, planting, and harvesting schedules, supporting food security and market stability \citep{ceglar_seasonal_2021}. In the energy sector, such forecasts help anticipate demand fluctuations, aiding in efficient resource allocation and grid management \citep{Dorrington_2020,buonocore_inefficient_2022,el-azab_seasonal_2025}. Additionally, governments and disaster preparedness agencies can use reliable forecasts to mitigate the impacts of extreme weather events, such as droughts or floods, thereby reducing economic losses and protecting vulnerable communities \citep{alfieri_glofas_2013,Samaniego_2019}.
Despite recent advances, forecasting European weather on long-term timescales remains challenging for both dynamical \citep{palmer_development_2004,mishra_multi-model_2019} and statistical methods \citep{Krikken_2024}.
While the weather forecast of physics-based \ac{NWP} is limited to about two weeks due to the intrinsic chaotic nature of the atmosphere \citep{lorenz_deterministic_1963}, statistical methods can explicitly leverage slowly varying, persistent, and more predictable components of the Earth system.
By Rossby wave propagation, these large-scale phenomena give rise to teleconnection patterns between distant regions of the Earth \citep{wallace_teleconnections_1981, ambrizzi_rossby_1995}, opening a window of opportunity to enlarge the forecast horizons.
In the northern hemisphere, the main mode of atmospheric variability, represented by the \ac{NAO} \citep{hurrell_overview_2003}, or the strongly related \ac{AO}, exerts influence over North America, Europe and part of Asia \citep{deser_role_2017, trigo_2002}, and is characterized by a typical persistence of 10 to 30 days \citep{feldstein_dynamics_2003, rennert_cross-frequency_2009, roberts_euro-atlantic_2023}.

The existence of persistent modes of the atmospheric state suggests that large-scale atmospheric circulation can be systematically decomposed into a finite number of dominant states \citep{Rossby_1940}.
Weather regimes (\ac{WR}), first introduced by \citet{rex_effect_1951}, are recurrent, quasi-stationary, and persistent states of atmospheric circulation \citep{hochman_atlantic-european_2021}. 
\ac{WR} may be defined separately for the summer and winter seasons, in which case four \ac{WR} are optimal in the Euro-Atlantic sector \cite{michelangeli_weather_1995}, or over the entire year, in which case seven \ac{WR} are optimal \cite{grams_balancing_2017}.
In the Euro-Atlantic sector, the \ac{WR} are strongly related to the \ac{NAO} \citep{michelangeli_weather_1995} and are known to exert considerable influence over the European weather \citep{van_der_wiel_influence_2019, trigo_circulation_2000, plaut_large-scale_2001, hall_north_2018, simpson_north_2024}\footnote{Some of the works mentioned here refer to different large-scale circulation patterns than \ac{WR}, but share similar ideas and comparable results.}.
Their persistent nature \citep[see e.g.][for a related discussion]{strommen_topological_2023, mukhin_metastability_2024}, often influenced by teleconnection patterns \citep{roberts_euro-atlantic_2023,michel_link_2011}, opens up the possibility of extending the forecast horizon in Europe beyond the limit posed by the chaotic nature of the atmosphere \citep{ferranti_flow-dependent_2015, ferranti_how_2018}. 
This justifies the increasing interest in assessing the capability of \ac{NWP} \citep{ferranti_how_2018, bueler_year-round_2021} and \ac{AI} models based on \ac{NN} architectures to forecast the transitions between large-scale circulation regimes in the sub-seasonal and seasonal timescales.
In the sub-seasonal scale, deep learning models such as convolutional neural networks (CNNs) and recurrent neural networks (RNNs) have been successfully applied to enhance large-scale circulation regimes by capturing complex spatial and temporal dependencies in reanalysis data \citep{chattopadhyay_predicting_2020, nielsen_forecasting_2022}. \citet{bommer_deep_2025} introduce advanced machine learning approaches to improve sub-seasonal predictions of the \ac{WR} by leveraging teleconnection patterns, further enhancing forecast reliability and skill. On the seasonal scale, \citet{hall_complex_2019} use complex modeling techniques to address the data-scarce conditions typical of this scale, aiming to forecast the first three principal components derived from \ac{EOF} analysis in the North Atlantic region.

However, while much research has focused on the evaluation of the forecast and on the impacts of the \ac{WR} on the European weather, the estimation of ground-level climate variables, such as temperature and precipitation, from Euro-Atlantic \ac{WR} remains largely unexplored and limited to linear methods, e.g. considering the mean surface weather variables associated with the dominant regime \citep{madonna_reconstructing_2021, roberts_euro-atlantic_2023}. Additionally, \citet{gerighausen_quantifying_2024} highlight the importance of intra-regime variability and suggest the use of continuous \ac{WR} indices to capture the regimes `subflavours' in place of the categorical classification commonly found in the literature \citep{madonna_reconstructing_2021, grams_balancing_2017, plaut_large-scale_2001, trigo_circulation_2000}.

In this study, we present two \ac{AI} models designed to reconstruct monthly anomalies in European temperature or precipitation, based on knowledge of the \ac{WR} indices. The models can capture and introduce complex non-linearities in the relationship between the \ac{WR} indices, which describe the state of the Euro-Atlantic atmospheric circulation, and the corresponding two-meter temperature and precipitation anomalies in Europe. 
After presenting the datasets (\refsec{sec:datasets}) and defining the \ac{WR} and the target fields (\refsec{sec:pre-processing}), we present the \ac{AI} model's architecture in \refsec{sec:model}. We evaluate its accuracy in reconstructing monthly mean temperature and total precipitation anomalies in winter and summer (\refsec{sec:reconstruction_skills}) and compare it with a standard baseline, namely the linear regression on individual \ac{WR} (\refsec{sec:baseline_comparison}). We then discuss the impact of changing the number of indices used to represent the atmospheric circulation (\refsec{sec:anom_recon_wr_num}). 
Given the potential future applications in a real forecasting scenario, we discuss the effects of introducing errors in the \ac{WR} indices, establishing a lower bound on the \ac{WR} prediction accuracy required to outperform the \ac{ECMWF} operational seasonal forecast system, SEAS5 (\refsec{sec:iwr_error}). In other words, we utilize the presented \ac{AI} models as a mapping from the \ac{WR} indices to the anomalies of interest to address the following questions:
\begin{itemize}
\item How does the model's reconstruction performance degrade as the number of input indices is reduced?
\item To what level of accuracy must the seven \ac{WR} indices be forecasted to outperform SEAS5 in predicting seasonal mean two-meter temperature and total precipitation?
\end{itemize}
We also test the accuracy of the \ac{AI} models in reconstructing the targeted anomalies using the SEAS5 forecast of the winters and summers \ac{WR} indices (\refsec{sec:iwr_seas5}) as a first evaluation of the models in a concrete forecasting scenario.

In summary, the \ac{AI} models presented here offer a promising approach to \ac{AI}-driven long-range weather forecasting by leveraging the predictability of large-scale atmospheric circulation and its relation with surface-level variables. 
The models demonstrate good reconstruction capability, as well as notable adaptability and stability in the presence of errors in the \ac{WR} indices, underlining their potential for seasonal forecasting applications. While they are applied in this study to reconstruct two-meter temperature and total precipitation anomalies on a monthly timescale, the architecture is sufficiently flexible to be adapted to other regions and variables without modification. Additionally, the model’s relatively lightweight design enables efficient training (requiring a single GPU), further supporting its practicality for broader implementation.

\section{Datasets}
\label{sec:datasets}

\subsection{ERA5 reanalysis}
ERA5, the fifth generation of \ac{ECMWF} global reanalysis \citep{Hersbach2020}, is the current state-of-the-art reanalysis and the most widely used data set in climate sciences. The ERA5 reanalysis spans years from 1940 onward, with a latency of 5 days and a Cartesian grid resolution of 0.25\textdegree, corresponding to approximately 31 km horizontal resolution at the equator. 
Using a bilinear interpolation, we upscaled the ERA5 data from the original 0.25\textdegree \texttimes 0.25\textdegree Cartesian grid to a 1\textdegree \texttimes 1\textdegree Cartesian grid.

The \ac{GPH} is one of the most widely used variables for studying teleconnections and decomposing atmospheric circulation into a set of dominant modes. We use the daily mean \ac{GPH} at 500 hPa (Z500) in the Euro-Atlantic region (80\textdegree W to 40\textdegree E, 30\textdegree N to 90\textdegree N) to compute the \ac{WR} and the related \ac{WR} indices (see \refsec{sec:weather_regimes}) used as input in our models.
The work aims to reconstruct the monthly mean two-meter temperature and total precipitation anomalies in Europe (i.e., within the domain 20\textdegree W to 30\textdegree E, 35\textdegree N to 70\textdegree N). We use the daily mean two-meter temperature and daily total precipitation to compute the respective daily anomalies, which are then aggregated monthly (see \refsec{sec:temp_prec_anom}).

\subsection{Numerical seasonal forecast: SEAS5}
\label{sec:seas5}

SEAS5 is the fifth-generation seasonal forecasting system developed by \ac{ECMWF} \citep{johnson_seas5_2019}. It provides global, long-range forecasts up to seven months ahead, using an advanced coupled ocean-atmosphere model. The SEAS5 forecast and hindcast of monthly aggregated \ac{GPH} at 500 hPa, two-meter temperature, and precipitation with 25 ensemble members were retrieved from the \ac{CDS} in the months from January 1981 to December 2024 on a 1\textdegree \texttimes 1\textdegree regular grid over the Euro-Atlantic region previously defined.

\section{Pre-processing}
\label{sec:pre-processing}
The following section details the pre-processing steps employed to derive the input and target variables for the \ac{AI} models. We then describe the models' architecture, training and evaluation in \refsec{sec:model}.

\subsection{Input of the AI-model: weather regimes indices}
\label{sec:weather_regimes}
\begin{figure*}
    \includegraphics{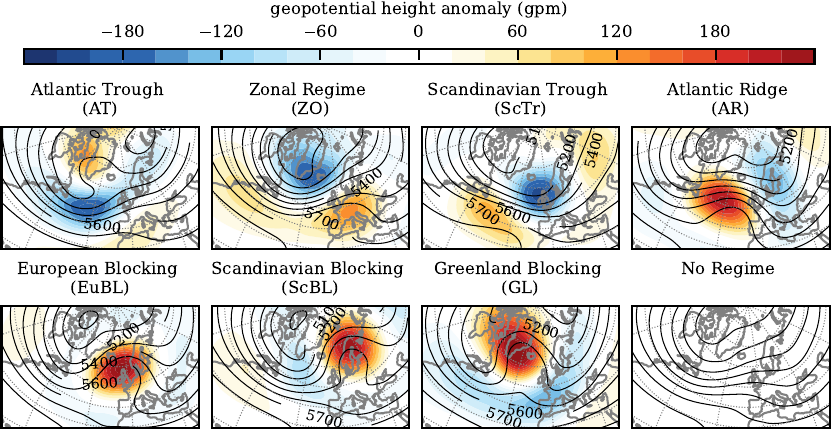}
    \caption{Cluster mean of the $Z500$ field (black isocontours) and corresponding average $Z500$ anomalies in winter (December, January, February), associated with the seven year-round Euro-Atlantic \ac{WR}. The average is performed in the 1981-2010 climatology. The ``No Regime'' refers to the mean atmospheric circulation when there is no \ac{WR} attribution according to the index life-cycle criteria \citep{grams_balancing_2017}.}
    \label{fig:winter_wr}
\end{figure*}

There are different methods of extracting low-frequency modes of variability of the atmospheric circulation \citep[refer to][for some examples]{hannachi_low-frequency_2017, schlef_atmospheric_2019, mukhin_revealing_2024, spuler_identifying_2024}. In this work, we use the method presented in \citet{grams_balancing_2017}, a natural extension of the procedure discussed in \citet{ferranti_flow-dependent_2015}.
The following briefly recaps the procedure for obtaining the seven year-round \ac{WR} \citep{grams_balancing_2017, bueler_year-round_2021}. 

The daily $Z500$ field is post-processed at each latitude $\phi$ and longitude $\lambda$ of the Euro-Atlantic sector to obtain the standardized anomaly field, $\tilde \Phi(t, \lambda, \phi)$:
\begin{enumerate}
    \item The climatology period is defined as the days from the first of January of 1981 to the end of 2010
    \item The calendar day climatology is obtained by applying a 15-day moving window centered on each calendar day ($\pm~7$ days) in the climatology period;
    \item The calendar day climatology is subtracted from the $Z500$ field, obtaining the $Z500$ anomalies, $\Phi(t, \lambda, \phi)$;
    \item The resulting field is 5 days low-pass filtered using a Savitzky–Golay filter\footnote{In agreement with \citet{robertson2020} and \citet{lee2023}, we did not find significant differences between the low-pass filtered and non-filtered \ac{WR}.};
    \item A calendar day normalization is computed by applying to the $Z500$ calendar standard deviation a 15-day moving window centered on each calendar day ($\pm~7$ days) in the climatology period, and then taking the spatial average on the full domain;
    \item The $Z500$ anomaly field, $\Phi(t, \lambda, \phi)$, is divided by the calendar day normalization.
\end{enumerate}
The last step is necessary to remove the seasonal cycle in the $Z500$ anomaly field amplitude, and it is required to define the year-round \ac{WR} \citep{grams_how_2020}.
\Ac{EOF} analysis \citep{hannachi_empirical_2007, xeofs_development_team} is performed on the standardized daily anomalies in the climatology period over the Euro-Atlantic domain. We retain the first seven \ac{EOF}, explaining $\approx 74 \%$ of the total variance.
K-means clustering is performed on the top seven principal components (PCs), yielding seven cluster centroids. 
The climatology mean of the $Z500$ field and $Z500$ anomalies
\footnote{although the clustering and the rest of the analysis are performed on the normalized anomalies, in the figure the non-normalized anomalies (i.e., the anomalies obtained in step 3) are shown for comparison with previous works.} 
associated with each cluster are shown in \reffig{fig:winter_wr}. The seven clusters mean correspond to the seven year-round Euro-Atlantic \ac{WR}, with three cyclonic regimes: \ac{AT}, \ac{ZO}, and \ac{ST}; and four blocking regimes: \ac{AR}, \ac{EuBL}, \ac{ScBL} and \ac{GL}. The ``No Regime'' refers to the mean atmospheric circulation when there is no \ac{WR} attribution: the attribution of each day to one of the \ac{WR} is subject to life-cycle criteria (see \cite{grams_balancing_2017} for the details).

The \ac{WR} index, $\Iwr$, is defined as the normalized projection of the daily standardized anomaly field, $\tilde \Phi(t, \lambda, \phi)$, onto each cluster mean field, $\tilde \Phi_{\rm wr}(\lambda, \phi)$, where wr = (\ac{AT}, \ac{ZO}, \ac{ST}, \ac{AR}, \ac{EuBL}, \ac{ScBL}, \ac{GL}):
\begin{equation} 
\label{eq:Pwr}
    I_{\rm wr}(t) = \frac{P_{\rm wr}(t) - \left \langle P_{\rm wr}(t) \right \rangle}{\sqrt{\left \langle \left( P_{\rm wr}(t) - \left \langle P_{\rm wr}(t) \right \rangle \right)^2 \right \rangle}}~,
\end{equation}
where the averages $\langle \cdot \rangle$ are performed in the climatology period, and $P_{\rm wr}$ is the aforementioned projection \citep[see][]{michelangeli_weather_1995}:
\begin{equation}
\label{eq:Iwr}
    P_{\rm wr}(t) = \frac{1}{\sum_{\lambda, \phi
    } \cos \phi} \sum_{\lambda, \phi} \tilde \Phi(t, \lambda, \phi) \tilde \Phi_{\rm wr}(\lambda, \phi) \cos \phi
\end{equation}

The method outlined here for deriving the seven \ac{WR} differs slightly from that used in \citet{grams_balancing_2017} and \citet{bueler_year-round_2021}. Key differences are detailed in \refapp{app:wr_differences}.

We are interested in quantifying the performance of the \ac{AI} models in reconstructing the targeted anomalies by changing the number of \ac{WR} used as input. Therefore, in addition to the seven weather regimes, we use the same procedure described before for the winter (December, January, February - DJF) and summer (June, July, August - JJA) seasons separately, changing the number of clusters to four. The four weather regimes obtained are the \ac{AT}, the \ac{ScBL}, the \ac{ZO}, and the \ac{GL} in winter (\reffig{fig:4clusters_winter_wr}), and in summer (\reffig{fig:4clusters_summer_wr}).
We also compute the \ac{NAO} index \citep{hurrell_overview_2003} separately for the summer and winter seasons, as the \ac{PC} of the first \ac{EOF}. The analysis is performed on the same Euro-Atlantic region and the same climatology period used to compute the \ac{WR}.

Finally, we aggregate the $\Iwr$ on a monthly basis. Indeed, we use the monthly average of the seven or four $\Iwr$ as an input variable in our \ac{AI} models (see \refsec{sec:model}).

\subsection{Perturbed indices}
As previously mentioned, it is valuable to assess the model's performance when using perturbed indices that simulate forecasting errors. This reflects a realistic scenario in which the predicted indices are not perfectly accurate.

Let us indicate with $\IwrP$ the set of indices with errors. The \ac{MAE} between $\Iwr$ and $\IwrP$ is a dimensionless number that depends on the normalization of the anomalies and the normalization of the \ac{WR} index itself. We then define the \ac{MARE} as $\langle |\Iwr - \IwrP| / |\Iwr| \rangle$, which quantifies the typical absolute relative error between the correct index $\Iwr$ and the perturbed index $\IwrP$.
We want to produce a perturbed index $\IwrP$ for a given value of the \ac{MARE}. For this purpose, let us define $\IwrP = (1 + f_{\rm wr}) \Iwr$, where $f_{\rm wr} \sim N(0, \sigma)$ are random variables sampled from a normal distribution with mean zero and standard deviation $\sigma$. It follows that the mean error is zero by construction. In other words, there is no bias in the $\IwrP$ error. The \ac{MARE} is $\langle |f_{\rm wr}| \rangle = \sqrt{2 / \pi} \sigma$. Fixing the distribution of $f_{\rm wr}$, that is, the value of $\sigma$, for a given value of the \ac{MARE} is straightforward. Also note that, since $f_{\rm wr}$ and $\Iwr$ are independent, the \ac{MAE} of the perturbed indices, $\langle | \Iwr - \IwrP| \rangle$, is simply given by $\rm{MARE} \times \langle | \Iwr | \rangle$. Unlike \ac{MARE}, the \ac{MAE}, therefore, depends directly on the statistical properties of the original indices.

For every value of the \ac{MARE} from 0 to 160\%, we compute \nreal random realization of $\IwrP$ sampling from $\Iwr \times N \left( 0, \sqrt{\pi / 2} ~{\rm{MARE} } \right)$. Each realization is then passed as an input to the \ac{AI} models (see \refsec{sec:model}) to obtain an estimate of the monthly average two-meter temperature and total precipitation in Europe for the months from January \testyearstart to December \RemoveSpaces{\testyearend}. 

\subsection{Target fields: temperature and precipitation anomalies}
\label{sec:temp_prec_anom}

The daily two-meter temperature and total precipitation anomalies are computed following the same procedure at points 1 to 3 described in \refsec{sec:weather_regimes}. Following the same pre-processing, the relevant relations between \ac{WR} and target anomalies are preserved, allowing the \ac{AI} models to exploit them for the reconstruction. Note that we do not normalize the anomalies by the calendar standard deviation, so we express the anomalies in Kelvin (K) for the monthly mean two-meter temperature and in centimeters (cm) for the monthly total precipitation.

The daily variables are then aggregated monthly: the daily mean two-meter temperature anomaly is averaged over every day in the month, while the daily total precipitation anomaly is summed. 
The monthly average two-meter temperature and total precipitation anomalies are the target variables that the \ac{AI} models are asked to reconstruct.

Note that we intentionally do not detrend the two-meter temperature data, allowing the \ac{AI} model to potentially capture the anthropogenic trend.

\subsection{SEAS5 bias-corrected forecast}
\label{sec:seas5-preprocessing}
Bias correction is a crucial step in calibrating the forecast by removing systematic biases. The distribution of simulated climate variables is adjusted to better match the observed historical data by mapping the quantiles of the model output to those of the reference dataset, ensuring that biases in the mean, variance, and higher-order statistical properties are minimized.
\Ac{EQM} is a widely employed method for bias-correcting the forecast data \citep{crespi_verification_2021, ratri_calibration_2021, golian_evaluating_2022}. 
We apply the \ac{EQM} to bias-correct monthly aggregated $Z500$, two-meter temperature, and precipitation for each of the 25 ensemble members and each of the seven forecasting lead times separately. 
To perform the \ac{EQM} we use the python package \texttt{xclim} \citep{bourgault_xclim_2023}.
The hindcast period, spanning the months from January 1981 to December 2010, is used to bias-correct the SEAS5 forecasts from January 2011 to December 2024, using ERA5 data upscaled to 1\textdegree \texttimes 1\textdegree resolution as reference. Note that the hindcast period used for the bias correction coincides with the climatological period.

The calendar day climatology and standard deviation of $Z500$, two-meter temperature and precipitation, derived from ERA5 data (see steps 1 to 5 in \refsec{sec:weather_regimes}), are aggregated to monthly values to obtain the SEAS5 monthly anomalies for these variables. Explicitly, the monthly average of the calendar day climatology (steps 1 to 4) is subtracted from the SEAS5 bias-corrected monthly forecasts. The resulting anomalies are then divided by the monthly average of the standard deviation computed in step 5.

Despite the \ac{AI} models are trained on the $\Iwr$ computed from ERA5 $Z500$ standardized anomalies, $\tilde \Phi(r,\lambda,\phi)$, it is interesting to evaluate its performance in reconstructing the monthly average two-meter temperature and total precipitation anomalies using the predicted $\Iwr$ computed from the bias-corrected SEAS5 $Z500$ forecast, $\IwrS$.
To obtain $\IwrS$ we project the forecast of the bias-corrected monthly mean $Z500$ standardized anomalies for each ensemble member $m$ and lead-time month $l$, $\tilde \Phi^{\rm SEAS5} (t, \lambda, \phi, l, n)$, to the seven mean cluster centers computed from the ERA5 daily anomalies, $\tilde \Phi_{\rm wr}(\lambda, \phi)$, (see \refsec{sec:weather_regimes}). The calculations follow closely \refeqnns{eq:Pwr} and \ref{eq:Iwr}, with the additional dimensions $l$ and $m$.

\section{AI-model}
\label{sec:model}
\begin{figure*}
    \centering
    \includegraphics[width=\linewidth]{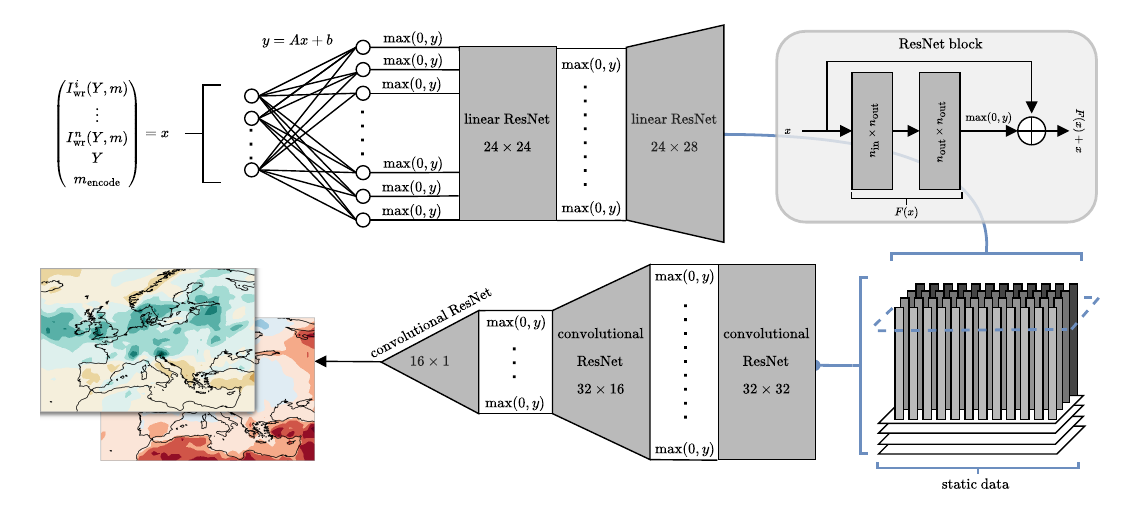}
    \caption{Schematic representation of the model's architecture. The input vector $x$, which includes the $n$ \ac{WR} indices, the year, and the encoded month, is passed as input to the model. After each layer, the output $y$ is passed to a ReLU activation function $\max(0,y)$. The $N = 28$ dimensional vector, output of the second linear ResNet block, is stacked pixel-by-pixel onto the static data (latitude and longitude grids, land-sea mask, and digital terrain model). The resulting $(N + 4) \times n_\lambda \times n_\phi$ tensor, where $n_\phi$ and $n_\lambda$ are the number of latitude and longitude grid points, respectively, is the input of the first convolutional ResNet layer. The number of features is then reduced by two by the second ResNet block and to one (temperature or precipitation field) by the final ResNet layer}.
    \label{fig:model}
\end{figure*}

The \ac{AI} models are designed to reconstruct the monthly average of the standardized two-meter temperature and total precipitation anomalies in Europe (\refsec{sec:temp_prec_anom}) from the monthly average \ac{WR} index, $\Iwr$ (\refsec{sec:weather_regimes}).

\subsection{Models' architecture and training}
\reffig{fig:model} schematically describes the models' architecture. The monthly average $\Iwr$, for each \ac{WR}, is passed along with the corresponding encoded month and year as input to the models. The month is encoded in twelve categories using one-hot encoding. One-hot encoding is a widely used technique in machine learning and data processing for representing categorical variables as numerical vectors. In our scenario, we encode the month = 1,...,12, into a twelve-dimensional vector, e.g., January (month = 1) is encoded in (1,0,...,0); February (month = 2) is encoded in (0,1,0, ...,0).
The inputs are then processed first by a linear layer and subsequently by a sequence of two ResNet blocks.
ResNet (Residual Network) introduces shortcut connections, allowing the model to learn residual mappings instead of direct transformations, which makes it easier to train deep architectures. The use of ResNet blocks helps mitigate the vanishing gradient problem, ensuring stable and efficient training of deeper networks, which is particularly beneficial when working with limited data.
The ResNet blocks take the 20-dimensional input (seven \ac{WR}, the year, and the twelve-dimensional encoded month) and yield a vector of $N = 28$ dimensions.
Expanding the feature space is a common strategy in \ac{NN}, as it enables the model to better capture and process the information in the input data by learning a more expressive representation of the temporal predictors.
The $N$-dimensional vector is stacked pixel-by-pixel with the two-dimensional static data: the latitude grid, the longitude grid, the land-sea mask, and the \ac{DTM} of the European region with the same resolution as the output two-dimensional field. 
Stacking these features pixel-by-pixel with the static spatial data enables the model to condition its reconstructions on both large-scale atmospheric drivers and local geographical features.
The resulting $(N + 4) \times n_\lambda \times n_\phi$ tensor, where $n_\phi$ and $n_\lambda$ are the numbers of latitude and longitude grid points, respectively, is the input of the following convolutional ResNet blocks, which reduce the number of features to one, i.e., to a two-dimensional field representing the target anomalies.
Convolutional layers capture both local spatial patterns and cross-grid dependencies, enabling the model to learn large-scale drivers and their geographically-modulated regional responses. 
We use 3x3 kernels in all the convolutional layers. These layers combine the information encoded in the feature vector with the static spatial data by selecting the components of the feature vector most relevant to each grid point. A 3\texttimes 3 kernel is sufficiently small to capture local structure.
We noticed that increasing the number of convolutional ResNet layers did not improve the performance of the model.
We employ ReLU activation functions to introduce non-linearity in a computationally efficient way, enabling the model to capture complex patterns.

The \ac{AI} models are trained separately to reconstruct two-meter temperature and total precipitation anomalies. They learn to approximate the target anomalies by adjusting their internal \nparams parameters to minimize the training loss between the predicted and observed anomaly fields. We choose the \ac{MSE} as training loss, defined in \refsec{sec:metrics}. For more details about the training setup and hyperparameters, the reader can refer to \refapp{app:training}.
To properly assess models' performance and their generalization ability, the dataset is divided into distinct training, validation, and test periods. 
The training set spans from January \trainyearstart to December \RemoveSpaces{\trainyearend}.
The months from \valstart to \valend are used to avoid overfitting, by halting the training when the validation \ac{MSE} reaches a minimum.
To finally assess the performance of the model, the remaining months from January \testyearstart to December \RemoveSpaces{\testyearend} are used. Note that the climatology period, used to compute the anomalies of the \ac{WR}, does not include the test dataset.

The models' performance can be slightly influenced by the order in which the input and output pairs of the train dataset are passed to the models during training. We use random ordering and repeat the training procedure \nseeds times to account for the dependence of the model's performance on the training data random ordering. 

The limited size of the training dataset (65 years) constrains the choice of models' architecture, preventing the use of larger and more complex architectures. Such architectures, with their higher number of parameters, would not only be computationally inefficient but also more prone to overfitting in data-scarce settings.

\subsection{Evaluation metrics}
\label{sec:metrics}

After performing an average on the \nseeds reconstructed anomalies by each of the \nseeds trained models, the quality of the reconstructed fields is evaluated using the metrics described in the following paragraphs.

\paragraph{Mean Absolute Error (MAE)} 
Is a commonly used metric to evaluate the accuracy of a model in predicting numerical values. It measures the average magnitude of errors by considering the absolute differences between actual and predicted values. Given a dataset with $n$ observations, where $y_i$ represents the actual values and $\hat y_i$ the predicted ones, \ac{MAE} is computed as
\begin{equation}\label{eq:mae}
    {\mathrm{MAE}} = \frac{1}{n} \sum_{i=1}^n \left | y_i - \hat y_i \right |~.
\end{equation}
A lower \ac{MAE} indicates a more accurate model. Since it relies on absolute differences, it treats overestimation and underestimation equally. 

\paragraph{Anomaly Correlation Coefficient (ACC)} 
ACC is a statistical measure to assess the similarity between predicted and observed anomalies in a dataset. It is commonly applied in climate science to evaluate the accuracy of deterministic forecasts. ACC quantifies how well the anomalies correlate between predictions and observations. Given a set of anomalies for $n$ observations, where $y_i$ represents the observed anomalies and $\hat y_i$ the predicted ones, \ac{ACC} is computed as
\begin{equation}\label{eq:acc}
    {\mathrm{ACC}} = \frac{\sum_{i=1}^{n} (y_i - \bar{y})\left( \hat{y}_i - \bar{\hat{y}} \right)}{\sqrt{\sum_{i=1}^{n} \left( y_i - \bar{y} \right)^2} \sqrt{\sum_{i=1}^{n} \left( \hat{y}_i - \bar{\hat{y}} \right)^2}}~,
\end{equation}
where $\bar y$ and $\bar{\hat y}$ denote the mean values of observed and predicted anomalies, respectively. The \ac{ACC} ranges from $-1$ to $1$, with values close to $1$ indicating a strong positive correlation, meaning that the model effectively captures the variations in anomalies. Conversely, values near $0$ suggest no correlation, while negative values indicate an inverse relationship between predicted and observed anomalies.

To assess whether an observed correlation is statistically significant, we compute the associated two-tailed p-value, derived from the exact distribution of the sample correlation coefficient, which follows a scaled beta distribution. The two-tailed p-value represents the probability of obtaining a higher or equal correlation assuming that there is no actual correlation (i.e., under the null hypothesis of zero correlation). In this context, a p-value below 0.05 is typically considered statistically significant, indicating that there is less than a 5\% probability that the observed correlation occurred by chance. Thus, correlations with $p < 0.05$ can be interpreted as significant at the 95\% confidence level.

\paragraph{Coefficient of efficiency (CE)} 
CE is a statistical measure used to evaluate the predictive skill of a model by comparing its performance to that of a validation climatology \citep{burger_verification_2007}, thus indicating the shared variance between the observed and estimated data throughout the verification period. Given $n$ observations, where $y_i$ represents the actual values, $\hat y_i$ the predicted values, and $\bar y$ the mean of observed values, the \ac{CE} is computed for each grid point as,
\begin{equation}\label{eq:ce}
    {\mathrm{CE}} = 1 - \frac{\sum_{i=1}^n (y_i - \hat y)^2}{\sum_{i=1}^N (y_i - \bar y)^2}~.
\end{equation}
A value of $\rm{CE} = 1$ indicates a perfect prediction, while $\rm{CE} = 0$ suggests that the model performs no better than the validation climatology. Negative values imply that the model performs worse than the reference. In this study, we use the climatology of the \testyearstart -- \testyearend testing period as validation climatology. When the \ac{CE} is evaluated during a specific season, only the related months are used to compute the validation climatology.

Since \ac{CE} is not bounded from below, it is possible to have negative outliers in its distribution. We then prefer to use medians instead of averages when aggregating the skills discussed in this paragraph to avoid the outliers from significantly imbalancing the results.

\vspace{5em}

To evaluate the aforementioned deterministic metrics on SEAS5, unless otherwise specified, we use the ensemble mean as the predicted value.

\paragraph{Continuous Ranked Probability Score (CRPS)}
CRPS is a widely used proper scoring rule for ensemble predictions \citep{Hersbach2020_crps}. The \ac{CRPS} measures the integrated squared difference between the cumulative distribution function (CDF) of the forecast ensemble, $F$, and the empirical CDF of the verifying observation, $H(\hat y) = \mathbf{1}_{\{\hat y \geq y\}}$, where $y$ is the observed value. Formally, the CRPS is defined as
\begin{equation}
    \mathrm{CRPS}(F, y) = \int_{-\infty}^{\infty} \left[ F(\hat y) - H(\hat y - y) \right]^2 \, d\hat y,
\end{equation}
which reduces to the mean absolute error (MAE) in the case of deterministic forecasts. Unlike the MAE, the CRPS accounts for the full ensemble distribution. Lower CRPS values indicate higher forecast skill, with a score of zero corresponding to a perfect forecast.
\newline\indent
In practice, when the forecast distribution is represented by a finite ensemble $\{ f_1, f_2, \ldots, f_m \}$ of size $m$, the CRPS can be approximated by
\begin{equation}
    \mathrm{CRPS}(F, y) \approx \frac{1}{m} \sum_{i=1}^{m} \left| f_i - y \right|
    - \frac{1}{2 m^2} \sum_{i=1}^{m} \sum_{j=1}^{m} \left| f_i - f_j \right|.
\end{equation}
The first term represents the mean absolute error of the ensemble members with respect to the observation, while the second term is a correction for the internal spread of the ensemble. 

\paragraph{Spread--Skill Ratio (SSR)}
The relationship between ensemble spread and forecast error is commonly assessed using the
\ac{SSR}. For an ensemble of size $m$ with mean forecast 
$\bar{\hat y}_t$ at time $t$ and verifying observation $y_t$, the ensemble spread is defined as
\begin{equation}
    \sigma_t = \sqrt{\frac{1}{m-1} \sum_{i=1}^m \left(\hat y_{i,t} - \bar{\hat y}_t\right)^2},
\end{equation}
while the forecast error of the ensemble mean is measured by the \ac{RMSE},
\begin{equation}
    \mathrm{RMSE} = \sqrt{\frac{1}{T} \sum_{t=1}^{T} \left(\bar{\hat y}_t - y_t\right)^2}.
\end{equation}
The \ac{SSR} is then given by the ratio between the temporal mean of the spread and the \ac{RMSE},
\begin{equation}
\mathrm{SSR} = \frac{\overline{\sigma_t}}{\mathrm{RMSE}}.
\end{equation}
An ideal ensemble forecast is characterized by $\mathrm{SSR} \approx 1$, indicating that the
predicted spread reliably reflects the actual forecast uncertainty. Values of $\mathrm{SSR} < 1$
denote an under-dispersive and overconfident ensemble, in which the spread underestimates the true forecast error, whereas $\mathrm{SSR} > 1$ indicates over-dispersion, with the ensemble spread being too large
relative to the error.

\section{Results}
\label{sec:results}

In the following sections, we evaluate the performance of the \ac{AI} models in reconstructing monthly mean two-meter temperature and total precipitation anomalies in winter (DJF) and summer (JJA) from January \testyearstart to December \RemoveSpaces{\testyearend}.
\refsec{sec:reconstruction_skills} presents the spatial reconstruction of these anomalies by the \ac{AI} models trained with the seven year-round \ac{WR} indices.
We compare the performance of the \ac{AI} model with the most commonly used reconstruction methods in \refsec{sec:baseline_comparison}.
We further investigate the skill of the models with respect to: (a) the number of \ac{WR} indices used for training and as input (\refsec{sec:anom_recon_wr_num}); (b) the impact of errors on $\Iwr$ on the reconstruction (\refsec{sec:iwr_error}); and (c) the use of $\IwrS$ as input (\refsec{sec:iwr_seas5}). The first two aspects address how the reconstruction quality degrades, identifying a threshold to achieve a better seasonal forecast than SEAS5. The third point examines the accuracy of the reconstructed anomalies when using \ac{WR} indices predicted by SEAS5. It can be interpreted as a hybrid \ac{NWP}-\ac{AI} framework for forecasting European mean two-meter temperature and total precipitation in winter and summer.

Although the \ac{AI} models reconstruct the anomalies over the entire European region as defined in \refsec{sec:temp_prec_anom}, in the following analysis, we restrict the evaluation to land areas only.

\subsection{Anomaly reconstruction with seven WR}
\label{sec:reconstruction_skills}
\begin{figure*}[t]
    \centering
    \includegraphics{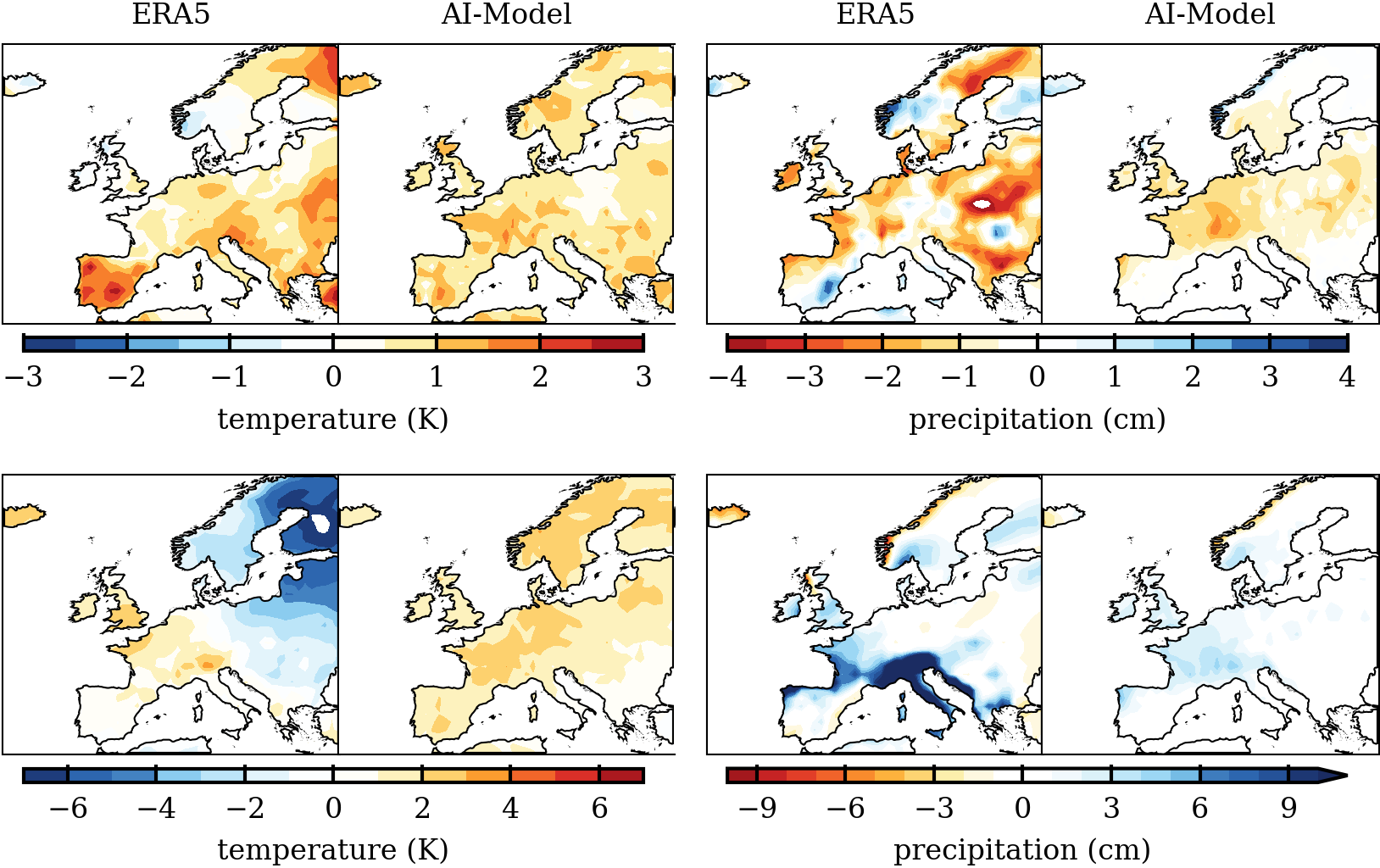}
    \caption{Qualitative comparison of reconstructed monthly mean two-meter temperature (left column) and total precipitation (right column) against ERA5. The top row displays the cases with the lowest mean squared error, while the bottom row shows those with the highest. The best (worst) reconstruction for temperature corresponds to July 2016 (February 2011), while for precipitation it is August 2013 (November 2019).}
    \label{fig:best_worst_reconstruction}
\end{figure*}

A qualitative comparison between the reconstructed monthly mean two-meter temperature (left column) and total precipitation (right column) against ERA5 is shown in \reffig{fig:best_worst_reconstruction}. The first row presents the best reconstruction, while the second row illustrates the worst reconstruction. Best and worst reconstructions are identified by the minimum and maximum \ac{MSE} on the test set, respectively, evaluated after training.
In the best-case scenario, the \ac{AI} model is able to capture part of the spatial distribution of monthly two-meter temperature and total precipitation anomalies. However, it generally fails to reproduce extremes and more localized patterns, resulting in reduced spatial variability compared to the reanalysis.
The worst-case scenarios usually occur when ERA5 anomalies are particularly strong. In such cases, the \ac{MSE} increases if the model does not adequately reproduce either the magnitude or the spatial distribution of the anomalies. A representative example is the monthly total precipitation: while the model successfully mimics the spatial distribution in parts of Central Europe, as well as the Iberian and Scandinavian Peninsulas, it underestimates the magnitude of the extremes, which in this case reach up to 20 cm.
The underestimation of extremes is a common issue for many \ac{NN} architectures trained using the \ac{MSE}.

\begin{figure*}[t]
    \centering
    \includegraphics{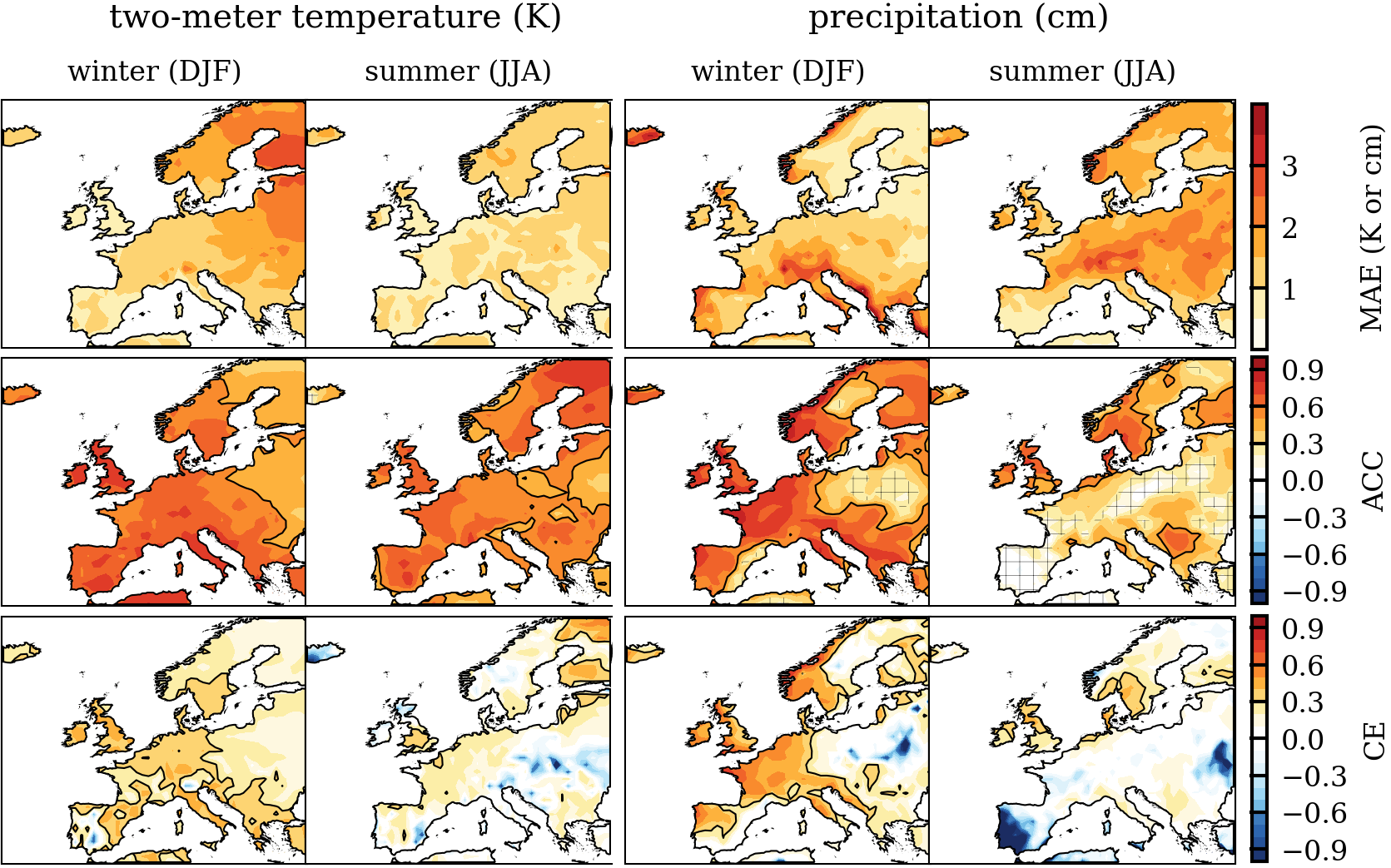}
    \caption{\ac{AI} models spatial \ac{MAE} (top), \ac{ACC} (middle) and \ac{CE} (bottom) in reconstructing the winters (first and third columns) and summers (second and fourth columns) mean two-meter temperature and total precipitation anomalies using seven \ac{WR} indices in the test years (from January \testyearstart to \RemoveSpaces{\testyearend}. Hatched areas mark regions with correlations that are not significant at $p > 0.05$. Black contours indicate regions with \ac{ACC} (\ac{CE}) higher than 0.5 (0.3).}
    \label{fig:temp_prec_skills}
\end{figure*}
\reffig{fig:temp_prec_skills} shows the spatial \ac{MAE}, \ac{ACC} and \ac{CE} in reconstructing the European monthly mean two-meter temperature and total precipitation anomaly using the seven year-round \ac{WR} indices, $\Iwr$, in the winters and summers during the test years. 

For the two-meter temperature anomaly, the \ac{MAE} maps indicate low reconstruction errors across most of Europe, especially in western, central, and southern regions, with values frequently below 1.5K. Errors are slightly higher in northern and eastern areas, particularly during winter, which is related to the higher variability and amplitude of the two-meter temperature anomalies in those regions, as demonstrated by the higher standard deviation (not shown).
The \ac{ACC} maps show high correlations (generally above 0.5) between reconstructed and observed anomalies, particularly over western, central, and southern Europe in both winter and summer, as well as in some parts of Northern Europe. This indicates the model effectively captures temporal variability in these areas.
The \ac{CE} metric partially mirrors this pattern, with values approaching 0.3 in winter, signifying that the model performs appreciably better than a climatological reference. In summer, the \ac{CE} remains low but generally above zero.

In the case of precipitation, the \ac{MAE} values are more spatially variable and generally higher than for temperature. Nonetheless, several regions, especially parts of Eastern and Northern Europe, exhibit relatively low \ac{MAE}, particularly in winter. 
The \ac{ACC} values for precipitation in winter are generally above 0.5 across most regions of Europe, except for some areas in central and eastern Europe. In the summer season, the \ac{ACC} values remain positive but are generally low, with large regions falling below 0.5 and correlations that are not statistically significant ($p > 0.05$). During this season, the \ac{AI} model driven by the \ac{WR} indices is largely unable to capture precipitation trends accurately, except in some regions of northern Europe.
\ac{CE} values for precipitation exhibit a similar spatial pattern of \ac{ACC}, with positive skill in central Europe and part of Scandinavia during winter but a marked decline in the central and eastern regions during summer.

Summer extratropical variability is less influenced by the large amplitude stationary waves and tropical forcing that characterize the boreal winter climate. As a result, the summer weather in the Northern Hemisphere is more influenced by regional-scale processes, such as mesoscale convective weather systems and land-atmosphere coupling \citep{ambrizzi_rossby_1995, Randal_2000, Siegfried_2011}.
This explains the lower skill reached by the \ac{AI} models in summer compared to winter in all the skills tested. In particular, summer \ac{CE} reveals that models using the Euro-Atlantic \ac{WR} are unable to accurately reconstruct the mean two-meter temperature and total precipitation anomalies in the Mediterranean and Black Sea regions. 
For these areas, the reconstruction could be improved by incorporating the Mediterranean large-scale circulation \citep{mastrantonas_extreme_2021} and including other predictors, such as soil moisture or rainfall deficit indices \citep{vautard_summertime_2007}. In fact, the authors of the latter study demonstrate that ``hot summers are preceded by winter rainfall deficits over Southern Europe''. Including indices that capture rainfall deficits (e.g., the Standardized Precipitation Index, SPI) and/or soil moisture (e.g., the Soil Moisture Deficit Index, SMDI) from the preceding season could enhance the reconstruction of the Southern European climate. We hypothesize that our model would be capable of capturing the relationship between these indices and the monthly summer anomalies of two-meter temperature and total precipitation in the region.

\subsection{Comparison with regressions on individual WR}
\label{sec:baseline_comparison}
We compare the performance of the AI models against a benchmark based on monthly composites of temperature and precipitation for each of the seven \ac{WR} \citep[see, e.g.,][]{grams_balancing_2017,madonna_reconstructing_2021,gerighausen_quantifying_2024}. We first identify the dominant WR for each month in the winters and summers from 1981 to 2010. A \ac{WR} is considered dominant if its \textit{monthly} index exceeds all other \ac{WR} indices and is larger than its standard deviation over the climatology.
We then compute the typical monthly two-meter temperature and total precipitation associated with each \ac{WR} by averaging the respective values over all months in which that \ac{WR} was dominant. Months without a clearly dominant \ac{WR} are assigned to a ``no-regime'' composite. Since each \ac{WR} can have a different impact on the ground variables depending on the season, the average is computed separately for the months in the winter (DJF) and summer (JJA) season. 
For the reconstruction period (\testyearstart onward), we assign to each month the composite corresponding to its active dominant \ac{WR}, or the no-regime composite if no dominant \ac{WR} could be identified, applying the same classification used in the climatology.
As for the \ac{AI} model, this approach relies on monthly average indices as predictors, enabling a direct comparison.

The composite reconstructions are evaluated using the deterministic metrics introduced in \ref{sec:metrics} against ERA5 monthly data. Relative scores are computed with respect to the AI-model as:
\begin{itemize} 
    \item relative improvement of the \ac{MAE}: (Composite MAE - AI-Model MAE) / Composite MAE,
    \item relative improvement of the \ac{ACC}: (AI-Model ACC - Composite ACC) / (1 - Composite ACC),
    \item relative improvement of the \ac{CE}: (AI-Model CE - Composite CE) / (1 - Composite CE).
\end{itemize}
These scores highlight the relative improvement of the \ac{AI} model with respect to the regressions on individual \ac{WR} as a benchmark.
\newline\indent
\begin{figure*}[t]
    \centering
    \includegraphics{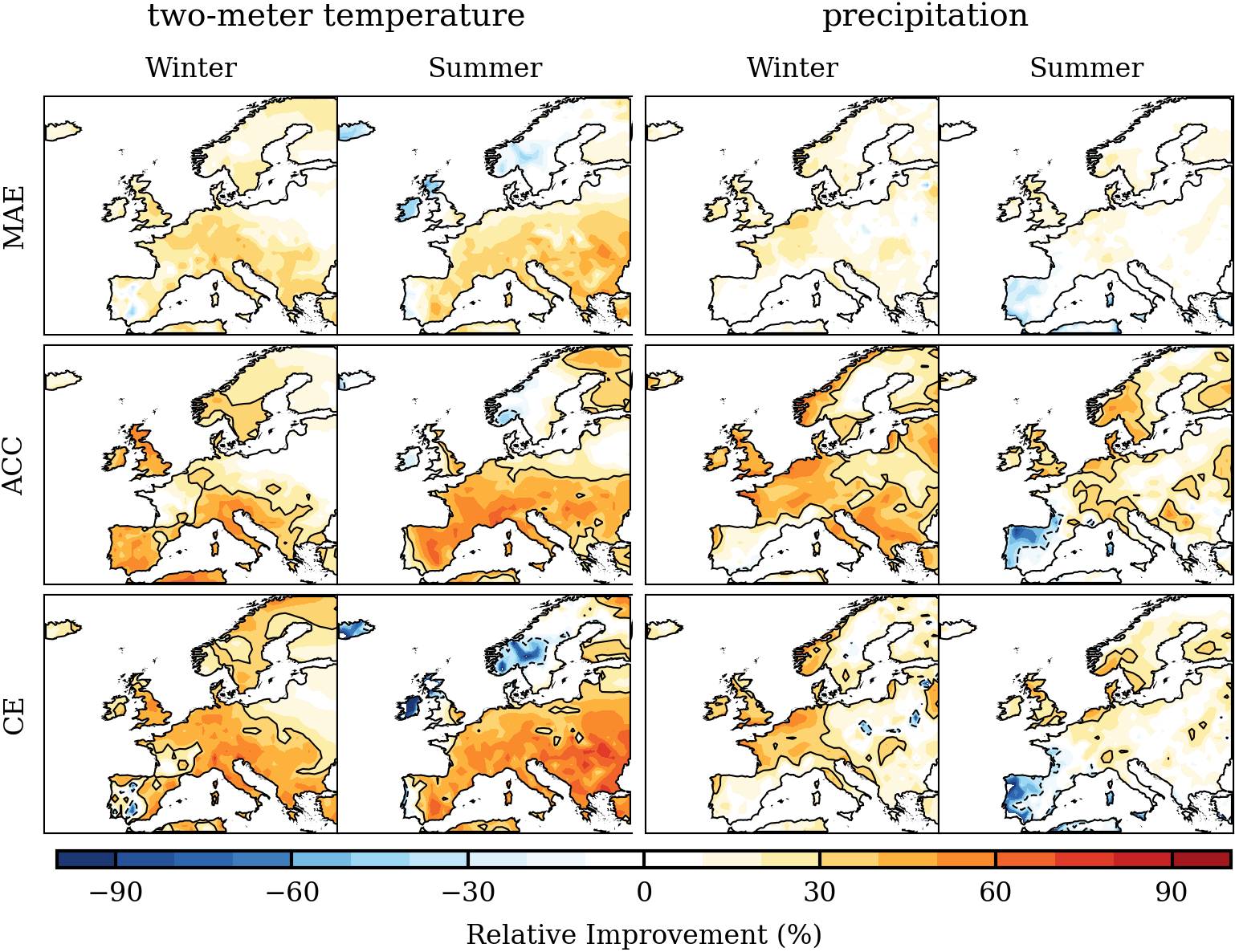}
    \caption{Relative improvement of the \ac{AI} models with respect to the regressions on individual \ac{WR} as benchmark. First (second) [third] row shows the relative improvement in \ac{MAE} (\ac{ACC}) [\ac{CE}] for both two-meter temperature and total precipitation separately for the winter (December, January, February) and summer (June, July, August) from \testyearstart to \RemoveSpaces{\testyearend}. Solid (dashed) contours indicate regions with relative improvement higher (lower) than 30\%.}
    \label{fig:composite_vs_model_7wr}
\end{figure*}
\reffig{fig:composite_vs_model_7wr} shows the relative improvement in percentage for both the two-meter temperature and total precipitation reconstruction in winter and summer from \testyearstart to \RemoveSpaces{\testyearend}. The \ac{AI} models outperform the simple composite benchmark across all seasons, variables, and metrics, with only a few regional exceptions. The improvement is particularly relevant for \ac{ACC} and \ac{CE} of the two-meter temperature reconstruction in central and Mediterranean Europe for both seasons, and particularly during summer, with the exception of some areas in Norway, Sweden, Greenland, and the British Isles. Also, the \ac{ACC} and \ac{CE} of the total precipitation show substantial improvements in many areas, but with less spatial consistency and with few regions characterized by negative scores in the Iberian peninsula.

Overall, the comparison demonstrates that while the composite method provides a simple and interpretable benchmark, the \ac{AI} models are capable of capturing subtle combinations of \ac{WR} patterns and reproducing regional variability more accurately. These results highlight the added value of data-driven reconstructions over traditional composite approaches, particularly when the underlying regime dynamics are complex.

\subsection{Impact of the number of WR on the reconstruction}
\label{sec:anom_recon_wr_num}

\begin{figure*}[t]
    \centering
    \includegraphics{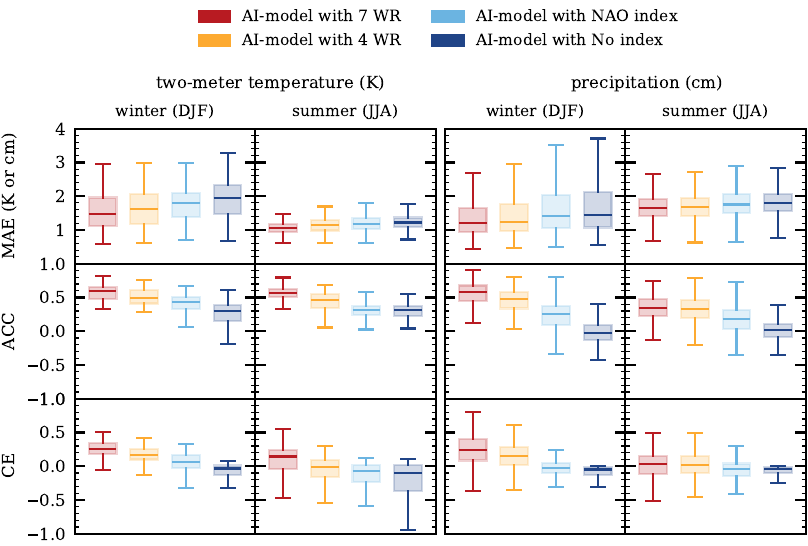}
    \caption{\testyearstart -- \testyearend winters (first and third columns) and summers (second and fourth columns) \ac{MAE} (top row), \ac{ACC} (middle row) and \ac{CE} (bottom row) of the two-meter temperature anomalies and precipitation anomalies, reconstructed by the \ac{AI} models trained with seven \ac{WR} (red), four \ac{WR} (orange), \ac{NAO} (light-blue) and zero (blue) indices. The box plots summarize the spatial data, showing the median (central line), interquartile range (box), and range.}
    \label{fig:7vs4WR}
\end{figure*}
In \reffig{fig:7vs4WR}, the reconstruction skills (\ac{MAE}, \ac{ACC}, and \ac{CE}) of the AI models for monthly mean two-meter temperature and total precipitation anomalies during winter (DJF) and summer (JJA) are compared across configurations using seven \ac{WR} indices (red), four \ac{WR} indices (orange), the \ac{NAO} index (light blue), and no indices (blue) as input. 
Analogously, \reftab{tab:temp_numWR} and \reftab{tab:prec_numWR} report the medians of the same metrics and the uncertainties on the last significant digit reported in parentheses. The uncertainties are computed as the semi-difference of the skill metrics’ medians across multiple model initializations with different random seeds, to quantify the spread due to the random seed initialization.
The models trained without indices (No index) consistently show the lowest scores for both temperature and precipitation. Despite this, the model retains limited skill for two-meter temperature, with median \ac{ACC} around $0.28 \pm 0.03$ in winter and $0.32 \pm 0.03$ in summer. This likely results from the AI model capturing the long-term climate trend through the year and month inputs. However, the spatial distribution of this correlation lacks statistical significance in large regions, as shown in \reffig{fig:temp_prec_skills_0wr}.
The inclusion of the \ac{NAO} index leads to some improvements. Since the \ac{NAO} alone explains approximately 25\% (16 \%) of the atmospheric circulation variance in winter (summer), it significantly enhances the \ac{ACC} for both variables in winter. Interestingly, for temperature in summer, the skills are comparable between the models using the \ac{NAO} index and without indices. We speculate that much of the reconstruction skill for the summer temperature likely arises from anthropogenic trends, with only a comparatively small contribution from the \ac{NAO}. A thorough assessment of this hypothesis is beyond the scope of the present work. The reconstruction of summer precipitation has instead a general improvement with respect to the model trained without indices.
Training the models with four \ac{WR} indices yields substantial gains over the NAO models across both seasons and variables. For temperature, the typical improvements in \ac{ACC} are about 28\% in winter and 16\% in summer, while for precipitation, they reach 74\% (DJF) and 100\% (JJA). The \ac{CE} also improves, particularly for the winter season. Although \ac{CE} remains near zero or slightly negative in summer, the four-index models exhibit far more dynamical skill and robustness than models with fewer indices.
Expanding to seven \ac{WR} indices results in reduced yet consistent improvements. The \ac{MAE} remains largely comparable to the four \ac{WR} setup, while \ac{ACC} increases by 27\% (DJF) and 35\% (JJA) for temperature and by 17\% (DJF) and 14\% (JJA) for precipitation. The \ac{CE} improves by approximately 113\% for temperature and 45\% for precipitation in winter, but remains low in summer.

In view of a data-driven forecast of \ac{WR} indices followed by the anomalies reconstruction, we suggest that using at least four \ac{WR} indices can provide an effective balance between reconstruction performance and index predictability, offering robust and meaningful reconstruction skills for both temperature and precipitation in winter and summer.

\begin{table}
    \centering
    \small
    \setlength{\tabcolsep}{4pt}
    \begin{tabular}{llcccc}
    \toprule
    \multicolumn{6}{c}{\textbf{Two-meter temperature}} \\
    \midrule
    Season & Metric & 7 WR & 4 WR & NAO & No idx \\
    \midrule
    \multirow{3}{*}{DJF}
        & MAE (K) & 0.78(3) & 0.91(5) & 1.04(6) & 1.01(2) \\
        & ACC & 0.65(4) & 0.51(1) & 0.40(2) & 0.28(3) \\
        & CE  & 0.32(7) & 0.15(2) & -0.01(4) & -0.04(1) \\
    \midrule\\
    \multirow{3}{*}{JJA}
        & MAE (K) & 0.77(2) & 0.88(6) & 0.93(3) & 0.82(2) \\
        & ACC & 0.50(5) & 0.37(3) & 0.32(1) & 0.32(3) \\
        & CE  & 0.07(5) & -0.03(6) & -0.09(3) & -0.08(2) \\
    \bottomrule
    \end{tabular}
    \caption{Median \ac{MAE}, \ac{ACC}, and \ac{CE} for monthly mean two-meter temperature in DJF and JJA using seven \ac{WR}, four \ac{WR}, \ac{NAO}, or no index as \ac{AI} model input. Uncertainties on the last significant digit (in parentheses) denote the semi-difference (half the difference between maximum and minimum) of the skill metrics’ medians across multiple model initializations with different random seeds.}
    \label{tab:temp_numWR}
\end{table}
\begin{table}
    \centering
    \small
    \setlength{\tabcolsep}{4pt}
    \begin{tabular}{llcccc}
    \toprule
    \multicolumn{6}{c}{\textbf{Total precipitation}} \\
    \midrule
    Season & Metric & 7 WR & 4 WR & NAO & No idx \\
    \midrule
    \multirow{3}{*}{DJF}
        & MAE (cm) & 1.34(1) & 1.41(3) & 1.66(2) & 1.86(1) \\
        & ACC & 0.63(1) & 0.54(2) & 0.31(2) & -0.02(2) \\
        & CE  & 0.32(2) & 0.22(2) & 0.01(2) & -0.05(1) \\
    \midrule\\
    \multirow{3}{*}{JJA}
        & MAE (cm) & 1.29(1) & 1.25(2) & 1.32(1) & 1.45(1) \\
        & ACC & 0.39(1) & 0.34(3) & 0.17(1) & 0.00(2) \\
        & CE  & 0.07(1) & 0.03(2) & -0.04(1) & -0.04(1) \\
    \bottomrule
    \end{tabular}
    \caption{As \reftab{tab:temp_numWR} but for monthly mean total precipitation in DJF and JJA using seven \ac{WR}, four \ac{WR}, \ac{NAO}, or no index as \ac{AI} model input. Uncertainties on the last significant digit (in parentheses) denote the semi-difference (half the difference between maximum and minimum) of the skill metrics’ medians across multiple model initializations with different random seeds.}
    \label{tab:prec_numWR}
\end{table}

\subsection{Reconstruction degradation with index errors}
\label{sec:iwr_error}

The \nreal perturbed indices for each value of the perturbation \ac{MARE} are used as input to the \ac{AI} models. The \ac{MAE}, the \ac{ACC}, and the \ac{CE} for the summer and winter seasons can be evaluated as usual for each of the \nreal reconstructed variables. After the skills are computed for each realization, we perform an average over the random realizations to evaluate the skills for the given \ac{MARE}.

\begin{figure}[H]
    \centering
    \includegraphics{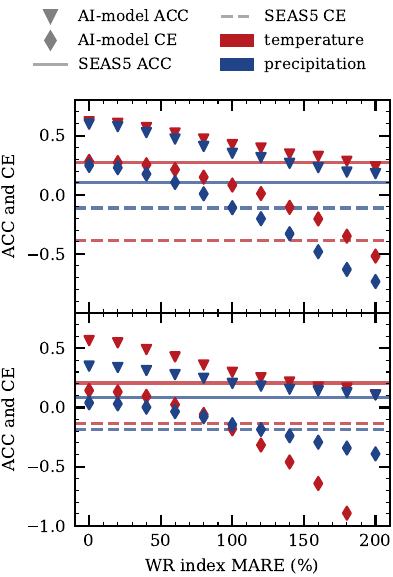}
    \caption{Spatial median of the mean two-meter temperature (red) and total precipitation (blue) \ac{ACC} (triangles) and \ac{CE} (rhomboids) reconstructed by the \ac{AI} models in the winter seasons (top panel) and summers (bottom panel) from \testyearstart to \testyearend, as a function of index \ac{MARE}. The values are averaged over \nreal random realizations of $\IwrP$ sampled from $\Iwr \times N \left( 0, \sqrt{\pi / 2} ~{\rm{MARE} } \right)$. Horizontal lines are the SEAS5 spatial median of the \ac{ACC} (solid) and CE (dashed) for the same variables and seasons.}
    \label{fig:MARE_MAE_ACC_CE}
\end{figure}

The spatial median of the mean two-meter temperature and total precipitation \ac{ACC} and \ac{CE}, averaged over the \nreal random realizations of the perturbed indices, are shown in \reffig{fig:MARE_MAE_ACC_CE} for the \testyearstart -- \testyearend winters and summers.
Increasing the \ac{MARE} of the perturbed indices, the \ac{AI} models' skills naturally degrade up to a point where they can be worse than the SEAS5 ensemble-mean \ac{ACC} and \ac{CE}.
Both in winter and summer, the median \ac{ACC} of the reconstructed mean two-meter temperature and total precipitation is always better than the SEAS5 ensemble-mean forecast \ac{ACC}, indicating a strong efficacy of the \ac{AI} models in capture the correlation between the Euro-Atlantic \ac{WR} and European two-meter temperature and precipitation also in the presence of Gaussian distributed errors on the input indices.
As for the ACC, the median \ac{CE} of the mean two-meter temperature in winter and the precipitation in summer is always higher than the respective SEAS5 ensemble-mean median \ac{CE}.
More interesting is the case of the median \ac{CE} of the reconstructed total precipitation and mean two-meter temperature, which becomes comparable to the relative SEAS5 skill at \ac{MARE} = 100\%.
Taking the lower value of MARE of 80\% as a reference, this poses a lower bound on the forecast accuracy of the seven $\Iwr$ required to have a seasonal forecast of the mean two-meter temperature and total precipitation with higher deterministic skills than the bias-corrected SEAS5 ensemble mean forecast used in this work.
We can express the results obtained here in terms of \ac{MAE} of the perturbed indices, $\langle | \Iwr - \IwrP | \rangle = 0.35$ for \ac{MARE} of 80\%.

\subsection{AI-models in a concrete forecasting scenario}
\label{sec:iwr_seas5}

To test the \ac{AI} models in a real forecasting scenario, we now analyze the reconstruction of the monthly mean two-meter temperature and total precipitation when using the monthly mean seven \ac{WR} indices forecasted by the bias-corrected SEAS5, $\IwrS$ (see \refsec{sec:seas5}), as model inputs. Although the \ac{AI} models are not inherently designed for ensemble forecasting, since they are trained deterministically using the \ac{MSE} loss, an ensemble can nevertheless be constructed by driving the models with the seven $\IwrS$ forecasts from each SEAS5 ensemble member. This allows us to evaluate not only deterministic metrics based on the ensemble means, but also probabilistic scores such as the \ac{CRPS} and the \ac{SSR} for both the hybrid \ac{NWP}-\ac{AI} models and SEAS5.
In the following, the SEAS5 forecasts for the winters (summers) are initialized on the first of November (May) to predict December, January, and February (June, July, and August).

\begin{figure}
    \centering
    \includegraphics{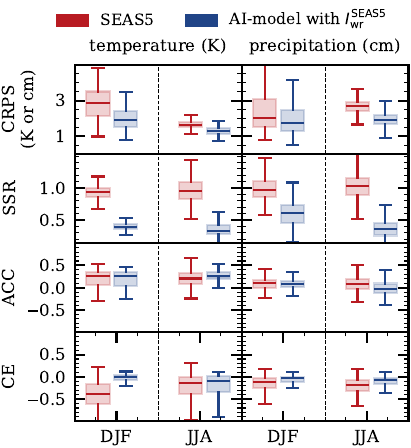}
    \caption{\ac{CRPS} (first row), \ac{SSR} (second row), \ac{ACC} (third row) and \ac{CE} (last row) for the SEAS5 (red) and the \ac{AI} models using $\IwrS$ as input (blue). The first and second columns show the winter (left half) and summer (right half) skills for the two-meter temperature and precipitation, respectively. The box plots summarize the spatial data, showing the median (central line), interquartile range (box), and range.}
    \label{fig:IwrSEAS5}
\end{figure}
\reffig{fig:IwrSEAS5} presents a comparative assessment of the hybrid \ac{NWP}-\ac{AI} models and the SEAS5 forecasting system for boreal winter and summer. 
The \ac{NWP}-\ac{AI} models typically exhibit better or comparable skills to the SEAS5 forecast. In particular, they achieve lower \ac{CRPS} values, particularly for the winter mean two-meter temperature and summer total precipitation anomalies. This enhanced performance is likely attributable to the use of an \ac{MSE} loss during training, which suppresses variability in the models' output. In contrast, SEAS5 tends to produce forecasts with higher variability, often resulting in higher \ac{MAE} but with an \ac{SSR} near the unity. The overconfidence of the hybrid \ac{NWP}-\ac{AI} approach is its major drawback: the ensemble spread is markedly reduced by the \ac{AI} model, which is not adequately designed and trained to face a probabilistic forecast. In the conclusions, we outline possible directions for addressing this limitation.
The reduced \ac{MAE} of the \ac{NWP}-\ac{AI} models translates into higher values of the \ac{CE}, especially in cases where the discrepancy in \ac{MAE} relative to SEAS5 is substantial. Nevertheless, both SEAS5 and \ac{NWP}-\ac{AI} models yield \ac{CE} values that are often close to or below zero, indicating limited skill relative to the climatology of the test dataset. 
Regarding the \ac{ACC}, the \ac{NWP}-\ac{AI} models slightly outperform SEAS5 for the two-meter temperature in both seasons. For the total precipitation anomalies, the \ac{ACC} is comparable in winter and worse in summer.

Considering the stability of the \ac{AT}-model in the presence of Gaussian-distributed errors in the \ac{WR} indices, as discussed in \refsec{sec:iwr_error}, the relatively poor performance of the \ac{NWP}-\ac{AI} models compared to the \ac{AI} model driven by perfect indices indicates that the SEAS5 forecast of the \ac{WR} (i.e., $Z500$) has limited skill. Indeed, the \ac{CRPS} of the indices degrades rapidly with forecast lead time, reaching around 0.6 already for forecasts of the month following the initialization (see \refapp{app:seas5_wr_skill}).

\section{Conclusions}
This study presents and evaluates two lightweight \ac{AI} based models for reconstructing monthly anomalies of two-meter temperature and total precipitation in Europe using Euro-Atlantic \ac{WR} indices. 
The models demonstrate robust skill in capturing both spatial and temporal variability of temperature, with somewhat reduced performance for precipitation. Reconstruction skills are higher during winter, reflecting the stronger influence of large-scale circulation patterns on surface climate during this season. 
We test the performance of the models by varying the number of \ac{WR} indices used as input. Despite the models using seven \ac{WR} achieving the best skills, four indices represent a reasonable trade-off between reconstruction accuracy and forecast feasibility.
We further assess the models' robustness to errors in the input \ac{WR} indices and identify a lower bound on their forecast accuracy necessary to outperform a state-of-the-art dynamical forecasting system, SEAS5. Specifically, to exceed SEAS5's skill, the indices of winter (DJF) and summer (JJA) must be predicted with an \ac{MARE} below approximately 80\%, corresponding to a \ac{MAE} of about 0.35 when the indices are normalized by the mean and standard deviation of the 1981 - 2011 climatology. 
Finally, in a realistic forecasting setup where the \ac{WR} indices are derived from bias-corrected SEAS5 forecasts, the hybrid \ac{NWP}-\ac{AI} models achieve comparable or superior performance relative to SEAS5, particularly for winter temperature and summer precipitation. These results highlight the potential of AI-driven approaches, grounded in dynamical predictors like \ac{WR} indices, to complement traditional numerical forecasting methods for improved long-range climate prediction.

Nonetheless, several limitations remain.
Training \ac{AI} models usually requires a large amount of data. We use monthly values from \trainyearstart to \trainyearend, totaling 852 points. The small training dataset forces the use of a lightweight architecture with a limited number of trainable parameters. 
Training with the \ac{MSE} loss encourages smooth reconstructions with reduced variability. While this leads to lower error scores, it also suppresses part of the natural variability of the climate system. This behavior is typical of a deterministic model trained on pointwise losses, which tend to regress toward the mean state rather than reproduce the full range of plausible outcomes. A natural extension would be to design the architecture for probabilistic reconstruction. For instance, an ensemble of reconstructions could be generated by training the network with proper probabilistic losses such as the \ac{CRPS}. This would allow the model to produce a distribution of possible realizations rather than a single deterministic trajectory, better capturing forecast uncertainty and enabling the use of ensemble-based verification metrics.

Despite these limitations, the approach is flexible and adaptable. The model can be extended to other regions, time scales, and surface variables. 
In particular, the sub-seasonal time scale presents a promising opportunity, offering both a richer dataset and potential for operational forecasting improvements.
Recent studies have proposed new approaches for computing the \ac{WR} \citep{geert_2025,spuler_identifying_2024,springer_unsupervised_2024,mukhin_revealing_2024}. Incorporating these \ac{WR} could potentially enhance the reconstruction performance of the proposed \ac{AI} model.

The next efforts will be directed towards a fully probabilistic \ac{AI}-driven method to forecast the \ac{WR} indices, which can be used as input to the \ac{AI} model presented here (or to a probabilistic extension) to obtain a forecast of the surface-level variables.

\section*{Acknowledgements}
\label{acknowledgements}
The authors thank Ilaria Tessari and Federico Fabiano for the valuable discussions on the year-round \ac{WR} calculations. We also thank Leonardo Aragao for the confrontation on the bias-correction of SEAS5.
This work was developed with financial support from ICSC–Centro Nazionale di Ricerca in High Performance Computing, Big Data and Quantum Computing, Spoke4 - Earth and Climate, funded by European Union – NextGenerationEU.

\section*{Data availability}
Daily post-processed ERA5 data on surface \citep{ERA5Copernicus_surface_daily} and pressure levels \citep{ERA5pressure_daily}, was generated using Copernicus Climate Change Service information 2020, obtained from the Copernicus Climate Change Service \citep{ERA5Copernicus_pressure_daily, ERA5Copernicus_surface_daily}. Neither the European Commission nor ECMWF is responsible for any use that may be made of the Copernicus information or data it contains.
The \ac{AI} models implemented in this work are written in PyTorch Lightning \citep{william_falcon_2020_3828935} and released under the MIT License through a GitHub repository (https://github.com/DSIP-FBK/anom\_recon) based on the Lightning-Hydra-Template \citep{Yadan2019Hydra}. The repository also contains all the scripts necessary to reproduce the results discussed in the paper (figures and tables).
The data used to train the models is available on Zenodo (https://zenodo.org/records/17276947).

\section*{Author contributions}
Alessandro Camilletti: Investigation, Conceptualization, Methodology, Data Curation, Formal Analysis, Software, Validation, Writing – Original Draft. Gabriele Franch: Methodology, Software, Writing – Review \& Editing, Resources, Supervision. Elena Tomasi: Supervision, Writing – Review \& Editing. Marco Cristoforetti: Funding acquisition, Project administration, Writing – Review \& Editing, Supervision. All authors participated in discussions throughout the development of the study and contributed to refining the manuscript. All authors have read and approved the final version of the manuscript.

\section*{Conflict of interest}
All authors declare that they have no conflicts of interest.

\bibliography{refs, extras}

\section*{Appendix}
\appendix{}
\renewcommand\thefigure{\thesection.\arabic{figure}}
\renewcommand\thetable{\thesection.\arabic{figure}}

\section{Four winter and summer WR}
\label{app:four_two_wr}
\setcounter{figure}{0}

\begin{figure}
    \centering
    \includegraphics{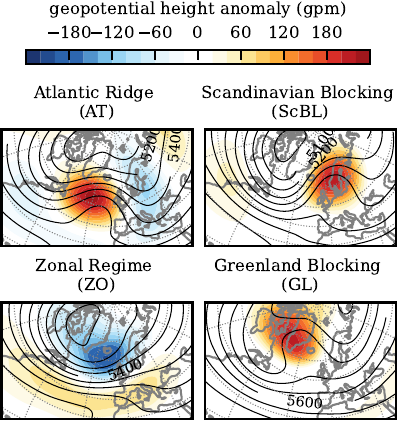}
    \caption{Cluster mean of the $Z500$ field (black isocontours) and corresponding average $Z500$ anomalies in winter (December, January, February), associated with the four Euro-Atlantic \ac{WR}. The average is performed in the 1981-2010 climatology.}
    \label{fig:4clusters_winter_wr}
\end{figure}

\begin{figure}
    \centering
    \includegraphics{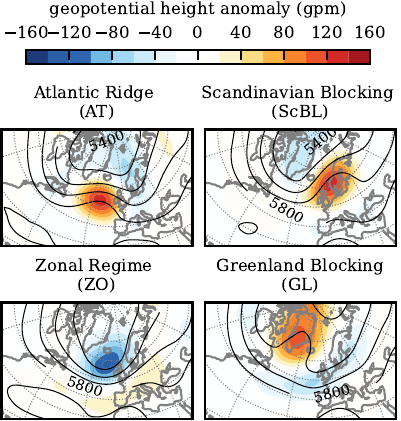}
    \caption{Cluster mean of the $Z500$ field (black isocontours) and corresponding average $Z500$ anomalies in summer (June, July, August), associated with the four Euro-Atlantic \ac{WR}. The average is performed in the 1981-2010 climatology.}
    \label{fig:4clusters_summer_wr}
\end{figure}

We compute the four \ac{WR} for winter and summer using the same methodology described in \refsec{sec:weather_regimes}, with the only change being the number of centroids used to cluster the data.
The mean $Z500$ field (black isocontours) and the corresponding average $Z500$ anomalies during winter (December, January, February) and summer (June, July, August), associated with the four Euro-Atlantic \ac{WR}, are presented in \reffig{fig:4clusters_winter_wr} and \reffig{fig:4clusters_summer_wr}, respectively.
The four \ac{WR} are the \ac{ZO}, \ac{ScBL}, \ac{AT} and \ac{GL}.

\section{Differences in the WR computations relative to Grams et al. (2017)}
\label{app:wr_differences}
\setcounter{figure}{0}

We now highlight the major differences to the procedure described in \citet{grams_balancing_2017} and in \citet{bueler_year-round_2021}. The authors of the mentioned work used the ERA-Interim reanalysis \citep{dee_era-interim_2011} six-hourly 500 hPa \ac{GPH} from 11 January 1971 to 31 December 2015; we instead use ERA5 daily 500 hPa \ac{GPH} from 1st of January \trainyearstart to 31 December \testyearend. Moreover, the anomalies computation, the \ac{EOF} analysis, the K-means clustering, and the definition of the cluster's mean field are performed only in the climatology period.
\cite{grams_balancing_2017} applied 90- and 30-day moving window averages to smooth the calendar day mean and standard deviation. In contrast, we use a 15-day centered moving average for each calendar day. We observed that using window sizes larger than approximately 30 days tends to preserve seasonal trends in the resulting anomalies.
This explains the minor difference that can be found when comparing \reffig{fig:winter_wr} with the corresponding Supplementary Figure 1 in \citet{grams_balancing_2017} and Figure 1 in \citet{bueler_year-round_2021}.

\section{Training setup and hyperparameters}
\label{app:training}
The models are trained using mean squared error (MSE) as the loss function (see \refsec{sec:metrics}). Training is performed using the Adam optimizer \citep{kingma2017adammethodstochasticoptimization} with mini-batches and early stopping based on the minimum validation loss, so the total number of epochs varies between runs and is not fixed \textit{a priori}.
We employ a learning-rate plateau scheduler: whenever the validation loss does not improve for 10 consecutive epochs, the learning rate is reduced by a factor of 0.5. The network includes dropout with a dropout rate of 0.2 to mitigate overfitting. Other training hyperparameters are reported in \reftab{tab:training_hyperparams}
Within reasonable ranges, hyperparameters such as batch size, initial learning rate, patience, and dropout rate have a limited impact on the final performance of the models. In our experiments, variations in these quantities result in changes comparable to the run-to-run variability due to randomness (see Tables 1 and 2).

\begin{table}[t]
    \centering
    \caption{Summary of training hyperparameters used in this work.}
    \label{tab:training_hyperparams}
    \begin{tabular}{ll}
        \toprule
        Hyperparameter      &Value \\
        \midrule
        Loss function                & \ac{MSE} \\
        Optimizer                    & Adam \\
        Initial learning rate        & $10^{-3}$ \\
        LR reduction factor          & 0.5 \\
        LR plateau patience          & 10 epochs \\
        Batch size                   & 8 \\
        Dropout rate                 & 0.2 \\
        \bottomrule
    \end{tabular}
\end{table}

\section{Anomaly reconstruction without indices}
\setcounter{figure}{0}
\begin{figure}
    \centering
    \includegraphics{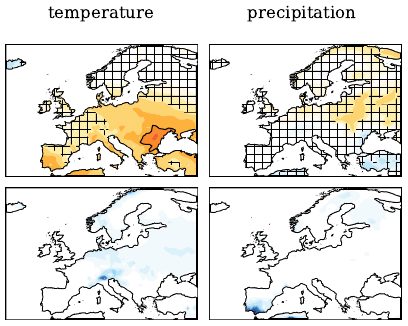}
    \caption{Spatial \ac{ACC} (top) and \ac{CE} (bottom) of the \ac{AI} model reconstruction of the winter mean two-meter temperature and total precipitation anomalies without using the information encoded in the \ac{WR} indices. The skills are computed in the test years (from January \testyearstart to \RemoveSpaces{\testyearend}. Hatched areas mark regions with correlations that are not significant at $p > 0.05$. Black contours indicate regions with \ac{ACC} higher than 0.5.}
    \label{fig:temp_prec_skills_0wr}
\end{figure}
As a baseline for evaluating the added value of the \ac{WR} indices, we trained the same \ac{AI} model described in \refsec{sec:model} to reconstruct monthly mean two-meter temperature and total precipitation anomalies without including the \ac{WR} indices as input. 
The obtained spatial \ac{ACC} and \ac{CE} for the winter mean two-meter temperature and total precipitation anomalies are shown in \reffig{fig:temp_prec_skills_0wr}.
Despite the absence of synoptic-scale information, the model exhibits low but positive \ac{ACC} in certain European regions, likely due to its ability to learn and exploit the underlying climate trend present in the training data. Nevertheless, the \ac{CE} value is close to zero or negative, indicating that the model is unable to replicate the amplitude of the fluctuations in the test dataset.

\section{Evaluation of the SEAS5 WR forecast}
\label{app:seas5_wr_skill}
\setcounter{figure}{0}

\begin{figure}
    \centering
    \includegraphics{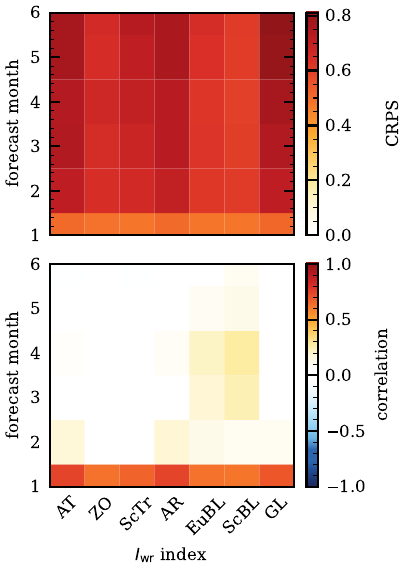}
    \caption{\ac{CRPS} (lower is better) and ensemble mean correlation (higher is better) of SEAS5 \ac{WR} indices forecast, $\IwrS$, with respect to ERA5 \ac{WR} indices, $\Iwr$. The labels refer to the \ac{WR} names: \ac{AT}, \ac{ZO}, \ac{ST}, \ac{AR}, \ac{EuBL}, \ac{ScBL}, \ac{GL}.}
    \label{fig:ERA5_SEAS5_indices_skills}
\end{figure}
As noted in \refsec{sec:iwr_seas5}, the low skills of the \ac{AI} model driven by the SEAS5 forecasted WR indices, $\IwrS$, suggest that the SEAS5 forecast of the \ac{WR} indices is not very skillful.
\reffig{fig:ERA5_SEAS5_indices_skills} show the \ac{CRPS} (lower is better) and the ensemble mean correlation (higher is better) between $\IwrS$ and $\Iwr$, i.e. between the monthly indices forecasted by the bias corrected SEAS5 and the indices computed from ERA5 Z500 (see \refsec{sec:weather_regimes} and \refsec{sec:seas5-preprocessing}).
Both the \ac{CRPS} and the ensemble mean correlation are good in the initialization month, but the skill degrades abruptly from the second forecasted month. As usually done in the literature, we initialize SEAS5 in November for winter forecasts (DJF), and in May for summer forecasts (JJA). The considered forecast is relative to the second, third, and fourth forecasted months, which, in general, has poor skills.

\end{document}